%% file: main.tex
\documentclass[letterpaper, 10 pt, conference]{ieeeconf}

\IEEEoverridecommandlockouts

\usepackage[utf8]{inputenc}

\usepackage{graphics} 
\usepackage{epsfig} 
\usepackage{dsfont}
\usepackage{amsmath} 
\usepackage{amssymb}  
\usepackage{multirow}
\usepackage{diagbox}
\usepackage{svg}
\usepackage{caption}
\usepackage{tikz}
\usepackage{subcaption}
\usepackage{booktabs}
\usepackage{adjustbox}
\usepackage{multirow}
\usepackage[ruled,vlined,linesnumbered]{algorithm2e}
\usepackage{float}

\usepackage[absolute,overlay]{textpos}
\usepackage{graphicx}
\usepackage{subcaption}
\usepackage{caption}

\newcommand{\OurMethod}{C-Free-Uniform}
\newcommand{\OurMethodMPPI}{CFU-MPPI}

\title{\LARGE \bf
C-Free-Uniform: A Map-Conditioned Trajectory Sampler for Model Predictive Path Integral Control
}

\author{Yukang Cao$^{1}$, Rahul Moorthy$^{1}$, O. Goktug Poyrazoglu$^{1}$ and Volkan Isler$^{2}$
\thanks{The authors are with the Robotics, Sensing and Networks Laboratory (RSN).
        $^{1}$University of Minnesota, Minneapolis, MN 55455, USA.
        $^{2}$The University of Texas at Austin, Austin, TX 78712, USA.
        Correspondence: {\tt\small cao00125@umn.edu}.}%
}
\begin{document}

\maketitle
\thispagestyle{empty}
\pagestyle{empty}

\begin{abstract}
\input{sections/abstract}
\end{abstract}

\section{Introduction}  \label{sec:introduction}
\input{sections/introduction}

\section{Related Work} \label{sec:related_work}
\input{sections/related_work}

\section{Preliminaries} \label{sec:preliminaries}
\input{sections/preliminaries}

\section{C-Free Uniformity Formulation} \label{sec:formulation}
\input{sections/formulation}

\section{\OurMethod{} Sampler} \label{sec:approach}
\input{sections/approach}

\section{Experiments} \label{sec:experiments}
\input{sections/experiments}

\section{Conclusion and Future Work} \label{sec:conclusion}
\input{sections/conclusion}

\bibliographystyle{IEEEtran}
\bibliography{citation.bib}
\end{document}

%% file: sections/abstract.tex
Trajectory sampling is a key component of sampling-based control mechanisms. Trajectory samplers rely on control input samplers, which generate control inputs $u$ from a distribution $p(u | x)$ where $x$ is the current state. 
We introduce the notion of Free Configuration  Space Uniformity (\OurMethod{} for short) which has two key features: (i) it generates a control input distribution so as to uniformly sample the free configuration space, and (ii)~in contrast to previously introduced trajectory sampling mechanisms where the distribution $p(u | x)$ is independent of the environment, \OurMethod{} is explicitly conditioned on the current local map. Next, we integrate this sampler into a new Model Predictive Path Integral (MPPI) Controller, \OurMethodMPPI{}. Experiments show that \OurMethodMPPI{} outperforms existing methods in terms of success rate in challenging navigation tasks in cluttered polygonal environments while requiring a much smaller sampling budget. 

%% file: sections/introduction.tex
\begin{figure}[t]
    \centering

    \includegraphics[width=\columnwidth]{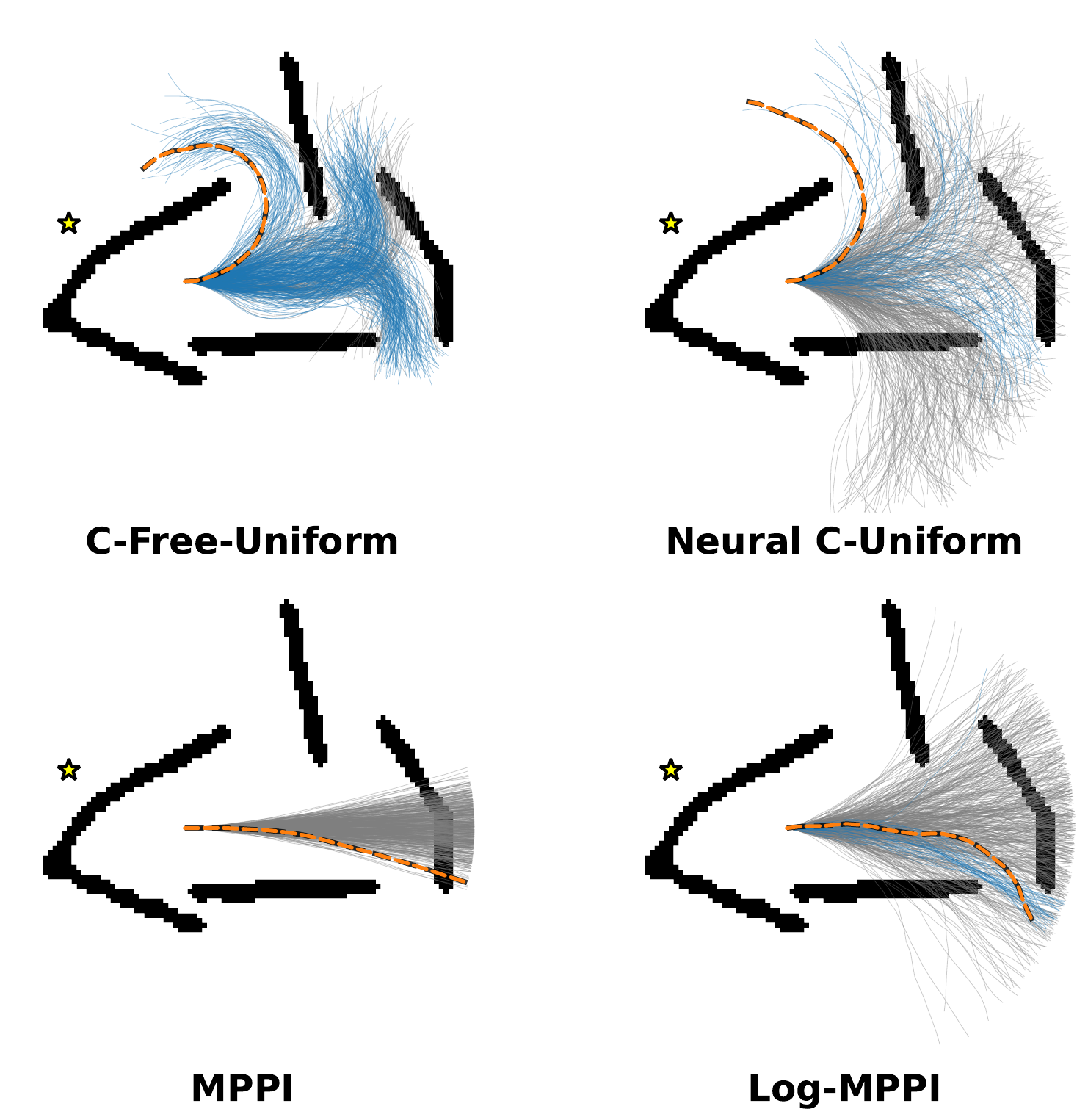}

    \caption{
        Comparison of trajectory sampling distributions (512 samples per method) in a cluttered environment, with obstacles in black, collision-free/colliding trajectories in blue/gray, the goal marked by a star, and the selected path in dashed orange. MPPI and Log-MPPI get stuck in a local minimum. While both C-Uniform and \OurMethod{} find a solution, \OurMethod{} shows a better path quality.
    }
    \label{fig:qualitative_comparison}
\end{figure}

Sampling-Based Model Predictive Control (SB-MPC) methods have emerged as powerful techniques for local trajectory planning. ~\cite{kazim2024recent}
The distribution used for sampling trajectories fundamentally affects a SB-MPC method's performance. ~\cite{hewing2020learning} 
A popular SB-MPC method, Model Predictive Path Integral (MPPI) ~\cite{williams2018information_MPPI}, relies on perturbing a nominal trajectory with zero-mean Gaussian noise.
This sampling strategy is highly sensitive to the initial choice of the nominal trajectory, and as a result, it is prone to getting stuck in local minima. 
Figure \ref{fig:qualitative_comparison} illustrates such an instance in a cluttered environment. MPPI and Log-MPPI (a variant of MPPI which uses the Normal-LogNormal distribution to achieve sample diversity) both fail due to poor nominal trajectory initialization.

Significant recent research has been dedicated to designing more effective sampling strategies to avoid local minima.  Approaches like Log-MPPI~\cite{mohamed2022autonomous_LOG_MPPI} use heavy-tailed distributions to improve exploration. However, they still generate a unimodal distribution centered on the nominal trajectory. To address multi-modality and capture complex cost landscapes, researchers have explored techniques such as Gaussian Mixture Models~\cite{wang2021variational}, or non-parametric approaches like Stein Variational Gradient Descent~\cite{honda2024stein} to guide samples toward diverse, low-cost regions.  More recently, the concept of C-Uniformity ~\cite{poyrazoglu2024c} has been introduced as an objective for uniform trajectory distribution over the reachable level-sets of a system. The uniform distribution is attractive because it maximizes the exploration of each level set. 
A major limitation of these sampling strategies is that they are oblivious to the specific environment. For example, while the uniform sampling strategy maximizes exploration over the C-Space, in a cluttered environment with narrow passages, most of the samples would be wasted as they would end up in collision states. See Figure~\ref{fig:qualitative_comparison}.

In this work, we introduce the notion of C-Free Uniformity, which aims to achieve a uniform distribution over the free portion of the configuration space (which we denote as C-Free). Our main contribution is a neural network model that takes a representation of the environment as input. The input can be the robot-centered map or the local sensing region around the robot, represented as an occupancy grid or Signed Distance Function (SDF). The network is trained to output a policy (action distribution), which then induces \OurMethod{} trajectories: they sample the free portion of the level sets uniformly. 


In summary, our contributions in this work are:
\begin{enumerate}
    \item We introduce the notion of \OurMethod, which aims to sample collision-free trajectories in the free configuration space uniformly. We present a learning-based method to compute the corresponding sampling policy in a supervised manner.

    \item We present the \OurMethodMPPI{}  controller which uses the learned sampler trained at a fixed velocity to achieve variable-speed navigation. To do so, we show how the trajectories can be scaled for various velocities and sampling rates. 

    \item We provide extensive experimental validation in simulation and on a physical robot. We demonstrate the robust generalization capabilities of the learned sampler across different operational parameters (velocity, time step, horizon) and the superior navigation performance of \OurMethodMPPI{} in complex environments. For instance, in challenging end-to-end navigation tasks, \OurMethodMPPI{} achieves a success rate of 57.4\% compared to 33.0\% for standard MPPI (with a 4096 sampling budget).

\end{enumerate}

%% file: sections/related_work.tex
Our work is at the intersection of sampling-based control and learning-based motion planning. The goal is to generate safe, environment-aware trajectory distributions.
\subsection{Trajectory Sampling Strategy in SB-MPC}
The performance of Sampling-Based Model Predictive Control (SB-MPC) is fundamentally tied to the underlying distribution of its samples. Traditional SB-MPC methods often rely on unimodal distributions, typically Gaussian. The Cross-Entropy Method ~\cite{rubinstein1999cross} iteratively fits a Gaussian distribution to the elite samples. Model Predictive Path Integral (MPPI)~\cite{williams2018information_MPPI} uses importance sampling to update a nominal trajectory based on Gaussian noise perturbation. Although effective in smooth cost landscapes, these methods are highly sensitive to initialization and prone to local minima. Variants like Log-MPPI~\cite{mohamed2022autonomous_LOG_MPPI} use heavy-tailed noise to improve exploration, and others focus on online covariance estimation to guide samples~\cite{yin2022trajectory}. To handle complex, multimodal cost landscapes, more expressive distributions are necessary. Gaussian Mixture Models are often used to capture multiple high-quality trajectory modes~\cite{wang2021variational}. Non-parametric approaches, such as Stein Variational Gradient Descent, use interacting particles to approximate complex distributions and guide samples toward low-cost regions~\cite{honda2024stein}.

In contrast to methods that optimize the distribution implicitly through the cost function, the concept of C-Uniformity~\cite{poyrazoglu2024c} introduces an explicit geometric objective: maximizing exploration by achieving a uniform distribution over the entire reachable space (Level Sets). While this maximizes exploration, it is environment-agnostic and wastes samples on trajectories that result in collisions.

\subsection{Safety in Sampling-Based Model Predictive Control}

In SB-MPC methods, such as MPPI ~\cite{williams2018information_MPPI}, the environmental constraints are typically handled by incorporating obstacle penalties. This approach does not fundamentally change the sampling distribution, which is inefficient in complex environments where the probability of sampling a feasible trajectory is low.

Significant research has focused on providing stronger safety guarantees.
Many approaches rely on reactive strategies. Control Barrier Functions (CBFs) are often integrated into MPC to filter unsafe actions~\cite{wang2024robust}. Methods like Shield-MPPI~\cite{yin2024chance} and Neural Shield-VIMPC (NS-VIMPC)~\cite{yin2025safe} utilize shielding or resampling strategies to correct or discard trajectories that violate safety constraints during the rollout process.

In contrast, our approach shapes the sampling distribution itself. C-Free Uniformity builds safety directly into the sampling process. We aim to generate distributions that are, by construction, uniform over the collision-free reachable space, rather than relying on penalties or corrective mechanisms. 

\subsection{Learning Map-Conditioned Sampler}
Learning-based methods have been increasingly used to improve the efficiency of motion planners. ~\cite{hewing2020learning} Several recent works focus on learning parameterized sampling distributions to enhance MPC performance. Some approaches use normalizing flows to learn generalizable trajectory distributions conditioned on environmental context~\cite{sacks2023learning},~\cite{power2024learning}. These methods often optimize the distribution implicitly through a task-based cost function. In contrast, we define an explicit target distribution (C-Free Uniformity) derived from network flow optimization and use a supervised learning approach to approximate it. This provides a clear objective for achieving both uniformity and safety.

To process spatial inputs for conditioning, U-Net architectures~\cite{ronneberger2015u} are widely used in robotics for retaining multi-scale information, often used to predict spatialized representations or cost maps~\cite{goel2022predicting}. We adopt this strategy, using a U-Net to generate a dense feature map from the SDF input, which is locally interpolated to condition the policy network.

%% file: sections/preliminaries.tex
We consider a robotic system with state $x_t \in \mathcal{X} \subseteq \mathbb{R}^p$ and control $u_t \in \mathcal{U} \subseteq \mathbb{R}^q$. The robot operates in a workspace $\mathcal{W} = \mathbb{R}^2$. The robot's state evolves according to discrete-time dynamics $x_{t+1} = f(x_t, u_t)$. We assume the dynamics function $f$ is Lipschitz continuous.  A trajectory $\tau$ of length $T$  is specified by an initial state $x_0$ and a control sequence $U = (u_0, \dots, u_{T-1})$, represented by the sequence of states $\tau = (x_0, x_1, \dots, x_T)$ obtained by recursively applying $f$ with inputs $x_t$ and $u_t$. We assume that all control inputs are valid for all discrete time steps $t$. 

\subsection{Vehicle Dynamics Model}
We utilize the Kinematic Single-Track model to represent the mobile robot's motion. ~\cite{althoff2017commonroad} The state of the robot is defined as $x = [p_x, p_y, \theta, v]^T$, representing the position, heading, and velocity. The control inputs are $u = [a, \delta]^T$, representing acceleration and steering angle. The continuous-time dynamics are given by:
\vspace{-1mm}
\begin{equation*}
\dot{p_x} = v \cos(\theta), \quad \dot{p_y} = v \sin(\theta), \quad \dot{\theta} = \frac{v}{L} \tan(\delta), \quad \dot{v} = a
\end{equation*}
where $L$ is the wheelbase of the vehicle.

For the C-Free Uniform sampler training (Sec \ref{sec:network_architecture}), we simplify the model by assuming a fixed nominal velocity $V_{train}$. The state space is reduced to $(p_x, p_y, \theta)$ and the control input is restricted to the steering angle $\delta$. The full CFU-MPPI controller utilizes the complete model for variable-speed navigation.

\subsection{C-Uniform Trajectory Sampling}
The concept of C-Uniform, introduced by Poyrazoglu et al. ~\cite{poyrazoglu2024c}, is a fundamental building block of our work. 
Their work defines \textit{Level Set} $L_t$ as the set of states $x \in \mathcal{X}$ that are reachable from an initial state $x_0$ in exactly $t$ time steps.

\textit{C-Uniformity} is the property that the configurations in each Level Set $L_t$ are sampled uniformly at random. The objective is to compute the probability distribution of control actions that leads to C-Uniformity across the planning horizon. The original work presents a network-flow-based optimization method to compute the action probability distribution to achieve C-Uniformity for a general robotic system.

\subsection{Model Predictive Path Integral (MPPI)}
MPPI ~\cite{williams2018information_MPPI} is a sampling-based optimization algorithm for model predictive control. It maintains a nominal control sequence $U_{nominal}$. At each time step, MPPI generates $K$ perturbed control sequences by adding noise sampled from a Gaussian distribution $\Delta U_k \sim \mathcal{N}(0, \Sigma_U)$. These sequences are rolled out using the system dynamics $f$ to produce candidate trajectories $\tau_k$.

Each trajectory is evaluated using a cost function $C(\tau_k)$. The nominal control sequence is then updated as a weighted average of the samples, where weights are calculated based on the trajectory costs using an inverse temperature parameter $\lambda$:
\vspace{-2mm}
$$ \omega_k = \frac{\exp(-\frac{1}{\lambda} C(\tau_k))}{\sum_{j=1}^{K} \exp(-\frac{1}{\lambda} C(\tau_j))} $$
\vspace{-2mm}
$$ U_{nominal} \leftarrow U_{nominal} + \sum_{k=1}^{K} \omega_k \Delta U_k $$
The first action of the updated $U_{nominal}$ is executed, and the process repeats.

%% file: sections/formulation.tex
We define the environment geometry (map) as $M$, which specifies the obstacle region $\mathcal{O}(M) \subset \mathcal{W}$. Let $\mathcal{A}(x) \subset \mathcal{W}$ represent the subset of the workspace occupied by the robot at state $x \in \mathcal{X}$. The collision-free configuration space is then defined as:
\vspace{-2mm}
$$ \mathcal{X}_{free}(M) = \{ x \in \mathcal{X} \mid \mathcal{A}(x) \cap \mathcal{O}(M) = \emptyset \}$$
\vspace{-3mm}

We define the set of all possible trajectories of length $T$ starting from $x_0$ as $\mathcal T(x_0, T)$. The set of all possible collision-free trajectories of length $T$ starting from $x_0$, conditioned on the map $M$, is denoted as $\mathcal{T}_{safe}(x_0, T, M)$, formally: 
\vspace{-3mm}
\begin{equation}
\begin{split}
\mathcal{T}_{safe}(x_0, T, M) = \{ \tau \in \mathcal{T}(x_0, T) \mid \\ x_t \in \mathcal{X}_{free}(M) \quad \forall t=0, \dots, T \}
\end{split}
\end{equation}

We define the minimum time required to reach a state $x$ via a safe trajectory as:
\vspace{-1mm}
\begin{equation}
\begin{split}
    T_{min}(x) = \min \{ t \mid & \exists \tau = (x_0, \dots, x_T) \\
    & \in \mathcal{T}_{safe}(x_0, T, M) \text{ s.t. } x_t = x \}
\end{split}
\end{equation}
\textit{Safe Level Set} $L_{safe}(x_0, t, M)$ is then defined as the collection of states whose minimum arrival time is exactly $t$:
\begin{equation}
L_{safe}(x_0, t, M) = \{ x \mid T_{min}(x) = t \}
\end{equation}


Our objective is to learn a parameterized stochastic policy $\pi_\theta (\mathcal{U} | x_t, M_{local})$, referred to as the sampler. Here $M_{local}$ is the local observation of the environment at state $x_t$. This sampler produces a trajectory distribution $P_\theta(\tau|x_0, M_{local})$ and a marginal state distribution $P_{\theta, t}(x|M_{local})$. 

We aim to achieve the \textbf{C-Free Uniformity} property. C-Free Uniformity defines the target state distribution as a uniform distribution in each Safe Level Set $L_{safe}(x_0, t, M)$ for all $t$ and zero elsewhere. This means that for any subregion $A \subseteq L_{safe}(x_0, t, M)$, the fraction of sampled trajectories passing through $A$ at time $t$ should be proportional to the measure of that subregion $\mu(A)$. We use the Lebesgue measure $\mu(\cdot)$ to represent the volume of a set in the configuration space, as detailed in~\cite{poyrazoglu2024c}. Formally, the \OurMethod{} target probability measure for any measurable subset $A \subseteq L_{safe}(x_0, t, M)$ is defined as:
\begin{equation} \label{TargetDistribution}
    P^*_t(A|M) = \!
    \begin{cases}
        \frac{\mu(A \cap L_{safe}(x_0, t, M))}{\mu(L_{safe}(x_0, t, M))} & \text{\hspace{-0.9em}if } \mu(L_{safe}(x_0, t, M)) > 0 \\
        0 & \hspace{-0.9em} \text{otherwise}
    \end{cases}
\end{equation}
\vspace{-5mm}

The learning objective is to find the optimal parameter $\theta^*$ by minimizing the sum of the Kullback-Leibler (KL) divergences across the horizon $T$: 
\vspace{-3mm}
$$ \theta^* = \arg\min_{\theta} \sum_{t=0}^{T} D_{KL}\left( P^*_t(\cdot|M) \,||\, P_{\theta, t}(\cdot|M) \right) $$

%% file: sections/approach.tex
\begin{figure*}[t]
    \centering

    \includegraphics[width=\textwidth]{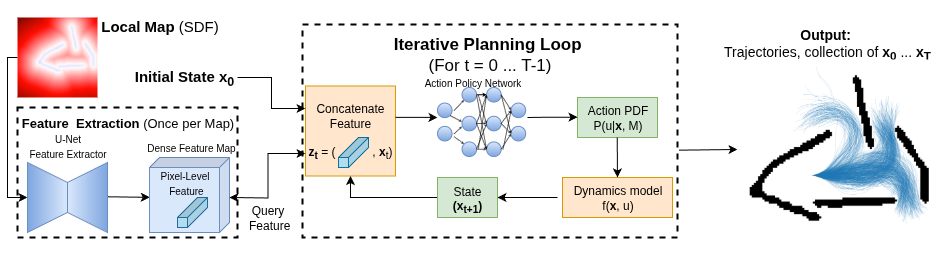}

    \caption{The \OurMethod{} System Architecture. The input is the robot's initial state and a signed distance function (SDF) representation of the environment. A U-Net extracts a dense feature map from the environment, which is used to condition a policy network inside an iterative planning loop. After training, the network outputs a control input probability distribution $p(u|x)$ which can be sampled to obtain uniform trajectories across the free space.
    In the diagram, blue components are neural networks, green are data representations, and orange are processing operations.
    }
    \label{fig:architecture_pipeline}
\end{figure*}

In this section, we present our approach to train a map-conditioned sampler that optimizes for C-Free Uniformity. 

\subsection{Supervised Learning Approach}
The existing unsupervised training approach ~\cite{poyrazoglu2025unsupervised} becomes computationally intractable when scaled to different map inputs. Its loss computation requires maintaining environment-specific level set representatives and involves operations that have quadratic memory complexity. Thus, we use a supervised learning approach to overcome this limitation. Our approach is to train a model $\pi_\theta$ to approximate the optimal \OurMethod{} action distribution $\pi^{*}$.

\subsection{\OurMethod{} Supervision Generation}
We adapt Algorithm 1 introduced in the C-Uniform paper ~\cite{poyrazoglu2024c} that uses a max-flow solver on a discretized state space to generate ground truth action probabilities. This max flow policy $\pi _{MF}$ is the surrogate distribution for the target policy $\pi^*$. Lemma 1 in ~\cite{poyrazoglu2024c} establishes that there exists a one-to-one correspondence between the max flow solution and the derived policy that leads to uniform distribution over the level sets. Therefore, training the model to mimic the $\pi _{MF}$ is a computationally tractable method to optimize for the C-Free Uniformity objective. 

The C-Uniform framework was modified to handle obstacles. The process is detailed in Algorithm \ref{algo:c_safe_pipeline}. First, we construct a collision-aware reachability graph to model the transitions between consecutive level sets. Next, inevitable collision states are pruned by ensuring that every node in the graph is reachable from the initial level set and the last level set. The remaining states represent the Safe Level Set. Lastly, we solve the max-flow optimization in these level sets to determine $\pi_{MF}$.

\vspace{-2mm}
\begin{algorithm}
\caption{C-Free Uniform (CFU)}
\label{algo:c_safe_pipeline}
\KwIn{
    \begin{tabular}[t]{@{}l}
        Map $M$ with obstacles $\mathcal{O}(M)$; Action set $\mathcal{U}$; \\ Dynamics $f$; Horizon length $N$; Time step $\Delta t$; \\ 
        Discretization resolution $\delta$;
    \end{tabular}
}
\KwOut{C-Free Level Sets $L_{safe}$ \& Max-Flow Policy $\pi_{MF}$;}

Construct the reachability graph $\mathcal{G}$ connecting discretized level sets $L[t]$ to $L[t+1]$, following the procedure in ~\cite{poyrazoglu2024c} (Algorithm 1, Lines 3-11).

\textbf{Modification:} During forward propagation, exclude any transition $(c_t, c_{t+1})$ if the resulting state $x_{t+1}$ is in immediate collision, i.e., $x_{t+1} \in \mathcal{O}(M)$.

We then prune locally safe states that inevitably lead to a collision within the horizon $N$. The remaining level set is the Safe Level Set $L_{safe}$


Compute the C-Free Uniform policy $\pi_{MF}$ by solving the Max-Flow optimization on the pruned graph $\mathcal{G}$ between consecutive Safe Level Sets, identical to the procedure in ~\cite{poyrazoglu2024c} (Algorithm 1, Lines 12-13).

\Return $L_{safe}, \pi_{MF}$\;
\end{algorithm}
\vspace{-3mm}

\subsection{Local Perception Data Pipeline} \label{sec:data_pipeline}
To train and evaluate a map-conditioned sampler, we require datasets containing perception inputs paired with the corresponding ground-truth \OurMethod{} policies ($\pi_{MF}$). We establish a standardized pipeline to generate these ``Local Perception Datasets" from any global environment.

\begin{figure}[thp]
    \captionsetup{aboveskip=3pt, belowskip=0pt} 
    \begin{center}
        \includegraphics[width=1.0\columnwidth]{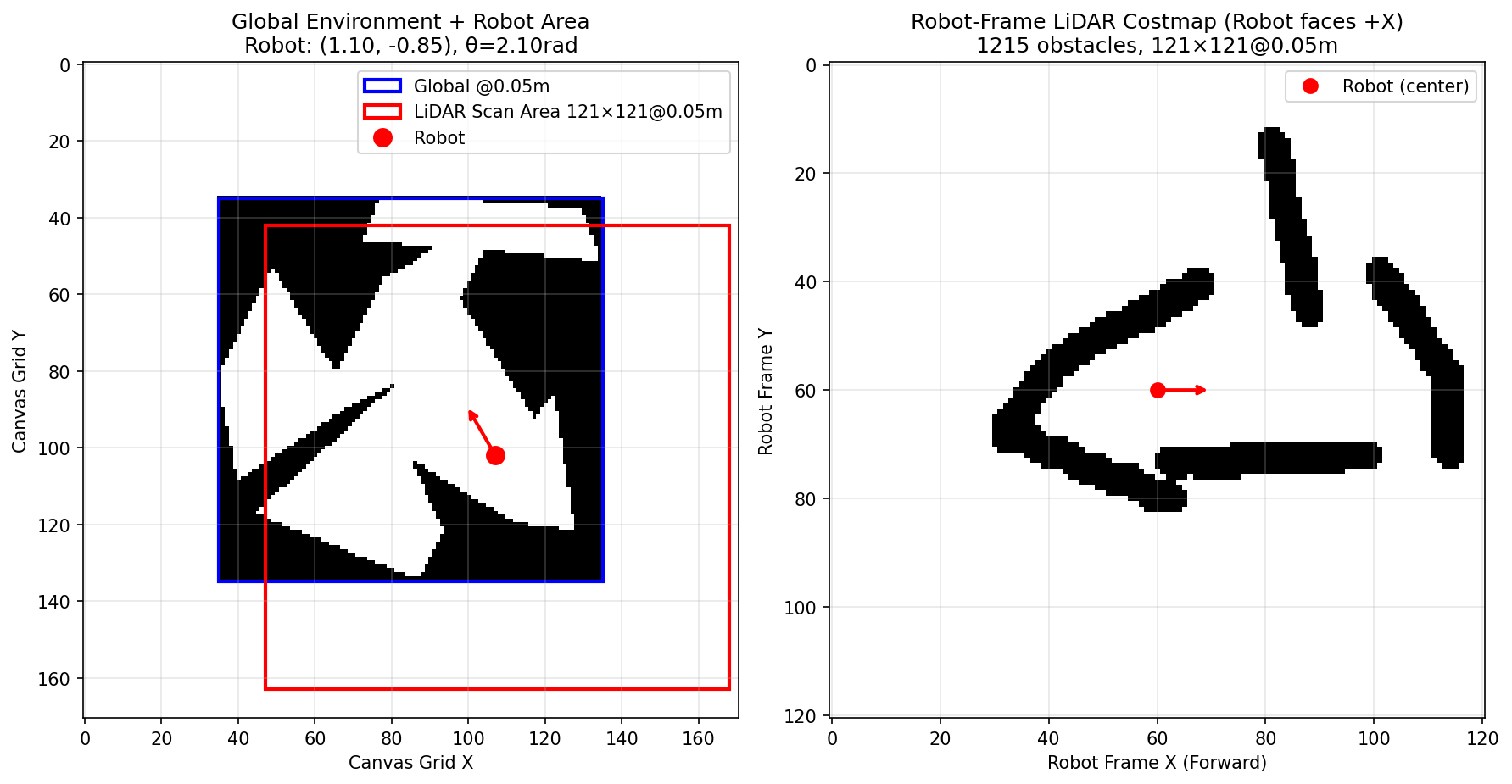}
    \end{center}
    \caption{Global map conversion to local perception map visualization. The complete map is shown on the left, and the extracted local map is shown on the right.}
    \label{fig:local_perception_pipeline}
\end{figure}

The pipeline first uniformly samples collision-free robot poses within a global environment. We simulate a 360-degree LiDAR scan for each pose to generate a robot-centered local perception map (e.g., SDF or costmap). Figure \ref{fig:local_perception_pipeline} visualizes this conversion. Finally, for a specific dynamic configuration (velocity $V$, time step $\Delta t$, horizon $T$), we compute the expert policy $\pi_{MF}$ and the Safe Level Sets using Algorithm \ref{algo:c_safe_pipeline}.

\subsection{Network Architecture and Training} \label{sec:network_architecture}
Our model consists of two components: a U-Net encoder-decoder that processes the input SDF map into a dense, pixel-level feature map, and an action policy network. The action policy network is a Multilayer Perceptron (MLP). To condition the policy on the map, we bilinearly interpolate the dense feature map to get the feature vector at the robot's $(x, y)$ location. This feature vector is then concatenated with the robot's state vector $x, y, \sin(\theta), \cos(\theta)$ and fed into the MLP (two hidden layers of size 2048 with ReLU activation and BatchNorm). The MLP outputs the action probability distribution over discretized steering actions. Figure \ref{fig:architecture_pipeline} shows a visualization of this process.

\textbf{Training Dataset:} We generated the training dataset using the pipeline described in Sec \ref{sec:data_pipeline}.   We sampled 5 poses from each environment in the BARN dataset ~\cite{perille2020benchmarking} (300 cluttered environments). Resulting in 1500 local perception scenarios (1200 training). The supervision $\pi_{MF}$ was computed using the nominal training dynamics with the velocity fixed at $2.5$ m/s, time step $\Delta t = 0.2$, and time Horizon $T=1.2s$.

\textbf{Training Protocol:} The entire model is trained end-to-end to minimize the KL divergence between its predicted action distribution and the ground-truth supervision $\pi_{MF}$. We use the Adam optimizer with a learning rate of $3\text{e-}4$ and weight decay of $1\text{e-}4$.

\subsection{\OurMethodMPPI{} Framework}
The generation of \OurMethod{} supervision (Algorithm \ref{algo:c_safe_pipeline}) relies on discretizing the configuration space to solve the Max-Flow optimization. The complexity of this process scales exponentially with the dimensionality of the configuration space. To maintain a tractable generation of computational data sets, the supervision $\pi_{MF}$ is computed using a fixed nominal velocity ($V_{train}$).

To navigate with variable velocity, we introduce the C-Free-Uniform Model Predictive Path Integral (\OurMethodMPPI{}) framework, which adopts the CU-MPPI framework from ~\cite{poyrazoglu2025unsupervised}. This controller first samples steering sequences using the robot's current velocity and then uses MPPI to refine the selected lowest-cost nominal action sequence in the full action space, which includes acceleration. The full \OurMethodMPPI{} control loop is detailed in algorithm \ref{algo:mc_cu_mppi}. 


\begin{algorithm}
\caption{\OurMethodMPPI{} Control Loop}
\label{algo:mc_cu_mppi}
\KwIn{\begin{tabular}[t]{@{}l}
        Current state $x_{curr}=(p_{curr}, v_{curr})$; Horizon $T$; \\ 
        Goal $x_{goal}$; Local Map $M_{local}$; Trained sampler $\pi_\theta$; \\ 
        Dynamics $f$; Cost $C$;
        Budgets $K_{init}, K_{mppi}$; \\
        MPPI parameters $\lambda$,  $N_{opt}$, $\Sigma_U$ (Noise covariance);
    \end{tabular}
}
\KwOut{Optimal control action $u_0^*$;}

Initialize pose batch $\mathbf{P}_0 \leftarrow \{p_{curr}\}^{K_{init}}$\;
\For{time step $t = 0$ to $T-1$}{
    $\mathbf{w}_t \sim \pi_\theta(\cdot | \mathbf{P}_{t}, M_{local})$\;

    $U_t \leftarrow$ Combine $v_{curr}$ with steering batch $\mathbf{w}_t$\;
    
    $\mathbf{P}_{t+1} \leftarrow f(\mathbf{P}_{t}, U_t)$\;
}

Resulting trajectories $\{\tau_{i}\}$, control sequences $\{U_i\}$\;
$S_i \leftarrow C(\tau_{i}, x_{goal}, M_{local})$ for $i=1 \dots K_{init}$\;
$i^* \leftarrow \arg\min_i S_i$\;
$U_{nominal} \leftarrow U_{i^*}$\;


\For{iteration $j = 1$ to $N_{opt}$}{
    $\{\Delta U_k\}_{k=1}^{K_{mppi}} \sim \mathcal{N}(0, \Sigma_U)$\;

    \For{$k = 1$ to $K_{mppi}$}{
        $U_k' \leftarrow U_{nominal} + \Delta U_k$\;
        $U_k' \leftarrow$ ClampControls($U_k'$)\;
        $\tau_k' \leftarrow$ Rollout using dynamics $f(x_{curr}, U_k')$\;
        $S_k' \leftarrow$ Calculate cost $C(\tau_k', x_{goal}, M_{local})$\;
    }

    Weights $\omega_k \leftarrow \text{softmax}(-S_k' / \lambda)$\;
    $U_{nominal} \leftarrow U_{nominal} + \sum_k \omega_k \Delta U_k$\;
}

\Return $u_0^* \leftarrow U_{nominal}[0]$\;
\end{algorithm}

%% file: sections/experiments.tex
To evaluate \OurMethod{} sampler and controller, we study the following questions through experiments:

\begin{enumerate}
    \item Can the map-conditioned sampler achieve a quantitatively higher degree of C-Free Uniformity than baselines? 
    \item How does the C-Free Uniformity change when the trained model is deployed under operational parameters different from those it was trained on?
    \item Does a higher degree of C-Free Uniformity translate into downstream performance, such as a higher success rate in single-frame planning and end-to-end navigation tasks when the sampling budget is constrained? 
\end{enumerate}

\subsection{Experiment Setup}
\textbf{Baselines: }
We compare our method against several sampling-based MPC baselines with different sampling distributions. We distinguish between the underlying \textbf{sampler} and the full \textbf{controller}. The sampler generates the initial trajectory distribution, and the controller uses this distribution to compute the final control actions. Our uniformity analysis and single-frame planning experiments analyze the properties of the sampler. The samplers selected include standard MPPI (Gaussian noise) ~\cite{williams2018information_MPPI}, Log-MPPI ~\cite{mohamed2022autonomous_LOG_MPPI} (normal-lognormal) noise distribution and the C-Uniform ~\cite{poyrazoglu2024c} sampler for uniform trajectory distribution. The selected controller baselines are MPPI, log-MPPI, and hybrid controllers, CU-MPPI ~\cite{poyrazoglu2025unsupervised}, and CU-LogMPPI.

For all MPPI variants, we implement a temperature parameter of $\lambda = 0.5$. A medium-level perturbation is selected, with a diagonal covariance matrix $\Sigma = \text{diag}(0.5^2, 0.1^2)$. The standard deviation entries correspond to acceleration and steering inputs, respectively. For hybrid CU-MPPI and \OurMethodMPPI, the sampling budget is split $50/50$ between respective initialization (C-Uniform or \OurMethod) and MPPI refinement stage.

\textbf{Trajectory Cost Function:}
The cost of a trajectory $\tau = (x_0, \dots, x_T)$ is designed to prioritize safety and goal-reaching behaviors. The total cost $C(\tau)$ is a weighted sum of distance costs, obstacle costs, and a terminal cost with an early-exit mechanism.
\begin{equation}
    C(\tau) = \sum_{t=0}^{T} \left( l_{dist}(x_t) + l_{obs}(x_t) \right) + l_{term}(x_T)
\end{equation}

\textit{Distance Cost} $l_{dist}$ penalizes the squared Euclidean distance to the goal $x_{goal}$:
$l_{dist}(x_t) = w_{dist} \cdot \|x_t[:2] - x_{goal}\|^2$.

\textit{Obstacle Cost} $l_{obs}$ is calculated based on the occupancy $O(x_t)$ of the robot's footprint within the local costmap. If the occupancy exceeds a predefined collision threshold $O_{thresh}$, a large penalty $P_{obs}$ is applied; otherwise, a cost scaled by the occupancy value is used to encourage clearance.

\textit{Terminal Cost} $l_{term}$ heavily weights the distance to the goal at the final state: $l_{term}(x_T) = w_{term} \cdot \|x_T[:2] - x_{goal}\|^2$.

\textit{Early Exit:} Cost accumulation stops immediately if a collision occurs ($O(x_t) > O_{thresh}$) or if the state enters the goal tolerance region. Subsequent costs for that trajectory are zeroed out, except for the collision penalty if applicable.

\textbf{Evaluation Environments and Datasets:}
We utilize two distinct setups for evaluation: for the end-to-end navigation tasks (Sec \ref{exp:end-to-end}), we use a collection of 300 complex, concave polygon environments from the Visdiff dataset ~\cite{moorthy2024visdiff}. This dataset is specifically designed to include topologically diverse scenarios that are challenging for local planners. To evaluate the sampler's trajectory distribution (Sec \ref{exp:uniformity}, \ref{exp:operational_scaling}, and \ref{exp:single_frame}), we generated Local Perception Datasets. These datasets were created using the standardized pipeline described in Sec \ref{sec:data_pipeline}, applied to the polygonal environments. 

\textbf{System Specification:} \label{system_spec} Simulation experiments were conducted on a platform equipped with a CPU Ryzen 5955WX and a GPU RTX 5090, running Ubuntu 25.04. Real experiments were conducted on a 1/5 scale custom mobile robot platform using Jetson AGX Orin for onboard computation. The robot uses the RPLIDAR S2 LiDAR as the perception module. The real-world experimental setup, with the mobile robot at its center, is shown in Fig. \ref{fig:real_world_setup}.

\subsection{Uniformity Analysis} \label{exp:uniformity}
To quantitatively evaluate the sampler's C-Free Uniformity property, we define a metric based on the KL divergence between the sampler's output state distribution and a smoothed version of the target distribution. The calculation is performed for each time step $t$ in the planning horizon:  we calculate the Safe Level Set using the algorithm \ref{algo:c_safe_pipeline}. The target distribution $P_t^*$ described in equation \ref{TargetDistribution} can be calculated by assigning a uniform probability to the discretized Safe Level Set. Next, for each sampler, we sample 50000 trajectories and obtain the empirical state distribution $Q_t$ by normalizing the count of states that fall inside each grid cell at time $t$. We add a small epsilon to the count of every bin in $P_t^*$ to ensure numerical stability; this creates $P_{smooth,t}^*$. The final C-Free Uniformity metric is: $D_{KL}(Q_t || P^*_{smooth, t})$.

To offer further insight, we also report two intuitive sub-metrics: the Collision-Free Ratio and the Entropy Ratio. The Entropy Ratio, defined as the empirical entropy divided by the maximum possible entropy, quantifies the uniformity of the safe samples, based on the maximum entropy principle~\cite{guiasu1985principle}.

Table \ref{tab:uniformity_summary} shows the average C-Free Uniformity across all level sets for the three metrics mentioned above. The \OurMethod{} sampler exhibits significantly lower KL divergence from the uniform distribution compared to baseline samplers, as it can maintain a high entropy ratio while showing 4-8 times more collision-free trajectories than baseline methods.
\begin{table}[h]
    \centering
    \caption{C-Free Uniformity Analysis}
    \label{tab:uniformity_summary}
    \begin{tabular}{lrrr}
    \toprule
    \textbf{Sampler} & 
    \textbf{\shortstack{Avg. KL \\ Div. $\downarrow$}} & 
    \textbf{\shortstack{Collision-Free \\ Ratio (\%) $\uparrow$}} & 
    \textbf{\shortstack{Avg. Entropy \\ Ratio (\%) $\uparrow$}} \\
    \midrule
    \textbf{\OurMethod} & \textbf{4.72} & \textbf{47.35}  & \textbf{90.73} \\
    C-Uniform & 8.67 & 11.81  & 90.63 \\
    MPPI & 11.74 & 5.62  & 37.66 \\
    Log-MPPI & 9.03 & 8.27  & 88.52 \\
    \bottomrule
    \end{tabular}
\end{table}

\subsection{Operational Parameter Scaling Analysis} \label{exp:operational_scaling}
A key feature of \OurMethodMPPI{} is the generalization capability of the learned sampler $\pi_\theta$ to different operational configurations from the one used during training. While the underlying vehicle dynamics model (Kinematic Single-Track) remains the same, operational parameters such as velocity ($V$), time discretization ($\Delta t$), and planning horizon ($T$) often vary during deployment and across robot systems.

For this experiment, we use the same trained model to predict the action probabilities for the robot state $(x, y, \theta)$. Then we propagate the sampled actions using a different set of operational parameters to generate the trajectories. This tests if the learned model generalizes to the changed Safe Level Set generated by the new parameters.

We measure the C-Free Uniformity using the average KL divergence metric established in Sec \ref{exp:uniformity} on the Local Perception Dataset. We test four categories of increasing difficulty:
(1) \textbf{Scale $V$} ($V = 1.25$ m/s); (2) \textbf{Scale $\Delta t$} ($\Delta t = 0.1$ s); (3) \textbf{Two-parameter scaling} ($V,\Delta t$); and (4) \textbf{Three-parameter scaling} ($V,\Delta t, T=2.4$ s).


For each configuration, we retrained the baseline C-Uniform model using the training parameters from ~\cite{poyrazoglu2025unsupervised}. Table \ref{tab:operational_scaling} shows the relative performance between different samplers. Note that KL divergence values are not directly comparable across each column because the target distribution is recomputed for each dynamic setting. This dynamic change alters the size of the Safe Level Set, so the scale of the KL divergence is changing. The results show that \OurMethod{} maintains a significantly closer distribution to the target compared to all baselines. An instance of this dynamic generalization can be seen in Figure \ref{fig:qualitative_comparison}, which shows a successful multi-modal distribution from the most challenging 3-parameter scaling test. 

\begin{table}[h]
    \centering
    \caption{
        Operational parameter scaling analysis of C-Free Uniformity ($D_{KL} \downarrow$). Nominal performance is in Table \ref{tab:uniformity_summary}.
    }
    \label{tab:operational_scaling}
    \begin{adjustbox}{max width=\columnwidth}
    \begin{tabular}{lcccc}
    \toprule
    \textbf{Sampler} & 
    \shortstack{\textbf{Scale $V$} \\ ($1.25$ m/s)} & 
    \shortstack{\textbf{Scale $\Delta t$} \\ ($0.1$ s)} & 
    \shortstack{\textbf{2-params} \\ ($V,\Delta t$)} & 
    \shortstack{\textbf{3-params} \\ ($V,\Delta t,T$)} \\
    \midrule
    \OurMethod  & \textbf{2.52} & \textbf{4.92}  & \textbf{2.57} & \textbf{7.33} \\
    C-Uniform   & 6.94 & 9.14                    & 6.34 & 9.90  \\
    MPPI        & 8.91 & 11.87                   & 8.17 & 11.81 \\
    Log-MPPI    & 6.99 & 9.36                    & 6.48 & 10.12 \\
    \bottomrule
    \end{tabular}
    \end{adjustbox}
\end{table}

\subsection{Single-Frame Planning Comparison} \label{exp:single_frame}
We conducted a single-frame planning experiment using the Local Perception dataset to evaluate each sampler's ability to find a feasible path to a goal within the planning horizon when the sampling budget is controlled. For each of the 300 local perception maps, we select a goal position uniformly at random within the last Safe Level Set. 10 trials per environment were performed, resulting in a total of 3000 trials for each sampling budget. Table \ref{tab:single_frame_planning} shows the average success rate across all environments. Success is defined as sampling at least one collision-free trajectory that reaches the goal position. The map-conditioned \OurMethod{} significantly outperforms the baseline methods across all selected sampling budgets. 

\vspace{-2mm}
\begin{table}[ht]
    \centering
    \caption{Success rates for single-frame planning}
    \label{tab:single_frame_planning}
    \begin{adjustbox}{max width=\columnwidth}
    \LARGE
    \begin{tabular}{c|cccc}
    \toprule
    
    \multirow{2}{*}{\textbf{\begin{tabular}[c]{@{}c@{}}Budget \\ (Trajectories)\end{tabular}}} & 
    \multicolumn{4}{c}{\textbf{Sampler Success Rate (\%) $\boldsymbol{\uparrow}$}} \\
    \cmidrule(lr){2-5}
    
    & \textbf{\shortstack{\OurMethod}} & \textbf{C-Uniform} & \textbf{MPPI} & \textbf{Log-MPPI} \\

    \midrule

    128  & \textbf{43.8} & 17.8 & 4.9 & 17.3 \\
    256  & \textbf{49.7} & 23.5 & 5.7 & 22.0 \\
    512  & \textbf{56.4} & 31.4 & 6.4 & 28.4 \\
    1024 & \textbf{62.8} & 37.8 & 7.3 & 33.3 \\
    2048 & \textbf{69.2} & 44.4 & 8.1 & 38.1 \\
    4096 & \textbf{74.8} & 50.8 & 9.0 & 43.1 \\
    
    \bottomrule
    \end{tabular}
    \end{adjustbox}
\end{table}
\vspace{-2mm}

\subsection{End-to-end Navigation} \label{exp:end-to-end}
To answer whether the higher degree of C-Free Uniformity improves the end-to-end navigation success rate, we evaluate map-conditioned hybrid controllers in the Polygon dataset. For each environment, a pair of positions is chosen to maximize the link diameter; they each serve as the start and the goal positions in 2 navigation tasks. Initial orientations are aligned with the shortest path from start to goal by running the A* search algorithm on a rasterized environment. For each task, 3 trials were performed, resulting in 1800 total simulation experiments for each specific controller configuration. 



We test the controller under two perception modules to analyze the robustness of the learned sampler
to different input types: (1) \textbf{Simulated LiDAR Perception:} A local costmap from a simulated 360-degree, 4.0m range LiDAR (Table \ref{e2e_navigation_success}). (2) \textbf{Oracle Local Perception:} A ground-truth local costmap, testing the performance ceiling (Table \ref{e2e_navigation_success_oracle_local_map}).

To isolate the contribution of the sampling distribution to the local planner's performance, no global path guidance is provided in both scenarios. This forces the controllers to rely entirely on their local perception and trajectory samples to navigate.

The average success rate is reported across all trials. Success is defined as navigating from start to goal position without collision. The results in both Table \ref{e2e_navigation_success} (LiDAR) and Table \ref{e2e_navigation_success_oracle_local_map} (Oracle) demonstrate a clear performance hierarchy across all selected sampling budgets. Our map-conditioned \OurMethod{} MPPI (\OurMethodMPPI) variants consistently outperform the unsupervised CU-MPPI variants, which in turn significantly outperform the standard MPPI and Log-MPPI baselines.

The relatively small gain when moving from simulated LiDAR to oracle local map input highlights the limitation of local planning in a complex and concave Polygon dataset. Perfect local perception does not resolve the issue of topological local minima like U-shaped obstacles. The consistent performance of the \OurMethodMPPI{} across both tables validates that the model generalizes to both realistic and idealized local map input.
\begin{table}[ht]
    \centering
    \caption{Success rates for end-to-end navigation in polygon environments with simulated LiDAR input.}
    \label{e2e_navigation_success}
    \begin{adjustbox}{max width=\columnwidth}
    \LARGE
    \begin{tabular}{c|cccccc}
    \toprule 
    
    \multirow{2}{*}{\textbf{\begin{tabular}[c]{@{}c@{}}Budget \\ (Trajectories)\end{tabular}}} & 
    \multicolumn{6}{c}{\textbf{Controller Success Rate (\%) $\boldsymbol{\uparrow}$}} \\
    \cmidrule(lr){2-7} 
    
    & \textbf{\begin{tabular}[c]{@{}c@{}} CFU \\ -MPPI\end{tabular}} & \textbf{\begin{tabular}[c]{@{}c@{}}CFU \\ -LogMPPI\end{tabular}} & \textbf{\begin{tabular}[c]{@{}c@{}}CU \\ -MPPI\end{tabular}} & \textbf{\begin{tabular}[c]{@{}c@{}}CU \\ -LogMPPI\end{tabular}} & \textbf{MPPI} & \textbf{\begin{tabular}[c]{@{}c@{}}Log \\ -MPPI\end{tabular}} \\

    \midrule 
    
    128  & 48.3 & \textbf{52.8} & 40.6 & 43.3 & 29.3 & 43.1 \\
    256  & 51.4 & \textbf{54.0} & 43.9 & 45.8 & 30.4 & 43.4 \\
    512  & 53.9 & \textbf{55.6} & 47.0 & 48.9 & 30.1 & 44.3 \\
    1024 & 54.5 & \textbf{54.7} & 49.5 & 49.8 & 30.7 & 45.4 \\
    2048 & 55.7 & \textbf{56.1} & 50.8 & 51.6 & 32.2 & 45.5 \\
    4096 & \textbf{57.4} & 56.4 & 52.6 & 52.1 & 33.0 & 46.7 \\
    
    \bottomrule 
    \end{tabular}
    \end{adjustbox}
\end{table}

\begin{table}[ht]
    \centering
    \caption{Success rates for end-to-end navigation in polygon environments with oracle local map input.}
    \label{e2e_navigation_success_oracle_local_map}
    \begin{adjustbox}{max width=\columnwidth}
    \LARGE
    \begin{tabular}{c|cccccc}
    \toprule 
    \multirow{2}{*}{\textbf{\begin{tabular}[c]{@{}c@{}}Budget \\ (Trajectories)\end{tabular}}} & 
    \multicolumn{6}{c}{\textbf{Controller Success Rate (\%) $\boldsymbol{\uparrow}$}} \\
    \cmidrule(lr){2-7} 
    & \textbf{\begin{tabular}[c]{@{}c@{}}CFU \\ -MPPI\end{tabular}} & \textbf{\begin{tabular}[c]{@{}c@{}}CFU \\ -LogMPPI\end{tabular}} & \textbf{\begin{tabular}[c]{@{}c@{}}CU \\ -MPPI\end{tabular}} & \textbf{\begin{tabular}[c]{@{}c@{}}CU \\ -LogMPPI\end{tabular}} & \textbf{MPPI} & \textbf{\begin{tabular}[c]{@{}c@{}}Log \\ -MPPI\end{tabular}} \\

    \midrule 
    
    128  & 49.2 & \textbf{53.6} & 42.8 & 48.1 & 30.7 & 44.1 \\
    256  & 50.3 & \textbf{55.2} & 46.7 & 48.8 & 31.8 & 44.7 \\
    512  & 55.4 & \textbf{57.3} & 49.8 & 50.5 & 31.9 & 45.2 \\
    1024 & 57.9 & \textbf{58.0} & 53.3 & 53.4 & 32.9 & 46.1 \\
    2048 & 57.8 & \textbf{57.6} & 54.1 & 55.0 & 34.3 & 46.3 \\
    4096 & \textbf{59.2} & 57.2 & 56.2 & 55.2 & 34.7 & 48.4 \\
    
    \bottomrule
    \end{tabular}
    \end{adjustbox}
\end{table}

\subsection{Real Experiments}
We validated the \OurMethodMPPI{} controller on a 1/5 scale custom Ackermann-steered mobile robot (Sec \ref{system_spec}), with state estimation provided by a Vicon motion capture system.

To isolate the performance of the local planner and to align it with the simulation experiment (Sec \ref{exp:end-to-end}), no global path guidance was provided to the controller. The robot does not have global map information. A fixed goal is set in the world frame, and the controller must reach it solely relying on its local perception and sampling distribution to make decisions.

We compare the performance of our \OurMethod{} Log-MPPI (CFU-LogMPPI) against the unsupervised C-Uniform Log-MPPI (CU-LogMPPI) and the standard Log-MPPI baseline, as Log-MPPI variants generally offer superior exploration capabilities. We evaluate performance across three sampling budgets: 128, 512, and 2048. 10 trials were conducted for each configuration. The primary metric is the Success Rate (SR\%), and the secondary metric is the Average Path Length for successful runs.

\begin{figure}[thp]
    \captionsetup{aboveskip=3pt, belowskip=0pt} 
    \begin{center}
    \includegraphics[width=0.75\columnwidth]{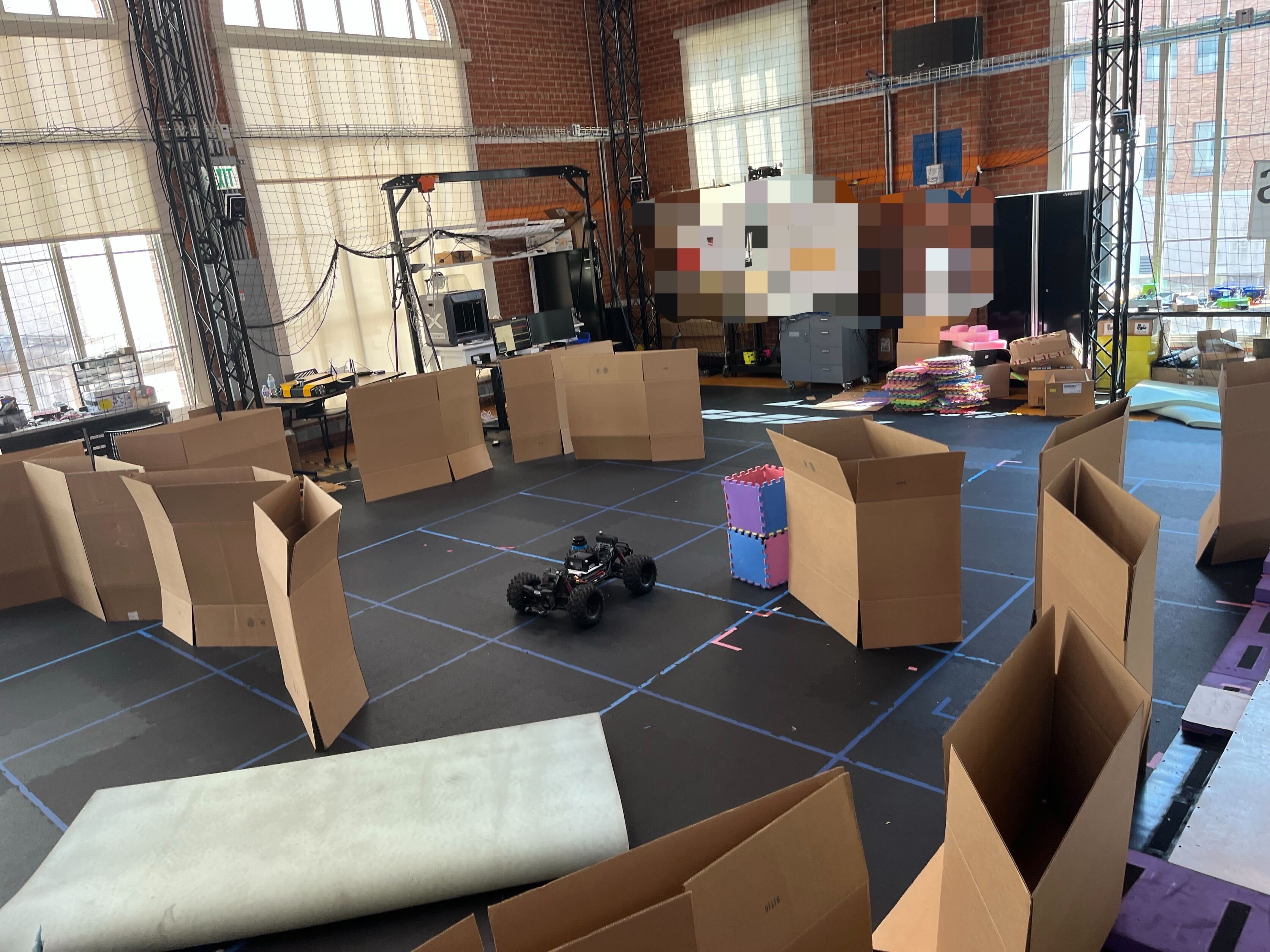}
    \end{center}
    \caption{The physical experiment setup. The robot navigated from a start position (bottom left) to a goal (top right) through a constrained environment requiring a severe S-turn. A foam pad on the ground is placed to generate perturbations.}
    \label{fig:real_world_setup}
\end{figure}

The experiment was conducted in an indoor environment shown in Figure \ref{fig:real_world_setup}. As summarized in Table \ref{tab:real_world_results}, our CFU-LogMPPI controller achieves the same success rate as CU-LogMPPI while producing higher-quality paths (shorter average length). Both controllers significantly outperform the standard Log-MPPI baseline across all tested sampling budgets.

\textbf{Computational Performance:}
We measured the inference time of the learned model on the Jetson AGX Orin using TensorRT optimization. The combined inference time for the U-Net feature extraction and the action policy network (processing a batch of 512 states) is approximately 4 ms. All controllers operate at 10 Hz.
\begin{table}[ht]
    \centering
    \caption{Success rates (SR\%) and average path length (meters) for real-world navigation in a constrained environment.}
    \label{tab:real_world_results}
    \begin{adjustbox}{max width=\columnwidth}
    \LARGE
    \begin{tabular}{c|cc|cc|cc}
    \toprule
    \multirow{3}{*}{\textbf{\begin{tabular}[c]{@{}c@{}}Budget \\ (Trajectories)\end{tabular}}} & 
    \multicolumn{6}{c}{\textbf{Controller Performance}} \\
    \cmidrule(lr){2-7} 
    & \multicolumn{2}{c|}{\textbf{CFU-LogMPPI (Ours)}} & \multicolumn{2}{c|}{\textbf{CU-LogMPPI}} & \multicolumn{2}{c}{\textbf{Log-MPPI}} \\
    & SR\% $\boldsymbol{\uparrow}$ & Length $\downarrow$ & SR\% $\boldsymbol{\uparrow}$ & Length $\downarrow$ & SR\% $\boldsymbol{\uparrow}$ & Length $\downarrow$ \\
    \midrule
    128  & \textbf{90} & \textbf{13.51} & \textbf{90} & 13.75 & 40 & 14.38 \\
    512  & \textbf{100} & \textbf{12.36} & \textbf{100} & 13.23 & 80 & 13.14 \\
    2048 & \textbf{90} & \textbf{13.07} & \textbf{90} & 13.25 & 60 & 14.83 \\

    
    \bottomrule
    \end{tabular}
    \end{adjustbox}
\end{table}

%% file: sections/conclusion.tex
In this work, we introduced the concept of C-Free Uniformity, a novel objective for trajectory sampling that prioritizes uniform exploration of the collision-free configuration space. We presented a trajectory sampler based on a map-conditioned neural network trained in a supervised fashion. The learned sampler was integrated into a new \OurMethodMPPI{} controller. We demonstrated how the sampler can be generalized to different operational speeds at runtime without the need for retraining.

Our experimental results demonstrate that the C-Free Uniform sampler achieves a significantly higher degree of uniformity over the safe reachable space compared to baselines, resulting in 4-8 times more collision-free samples. This improvement directly translates to superior downstream performance in challenging end-to-end navigation tasks and real-world scenarios.

One limitation of the current approach is that the trajectory generation process is iterative. The sampler outputs the action distribution for the current state only. To obtain the full trajectory, this process needs to be repeated for each level-set. Consequently, the inference time depends on the number of steps in the trajectory horizon. Future work could focus on improving the architecture to output the control-input probabilities for all level-sets in one forward pass. 